\documentclass[11pt,letterpaper]{article}
\usepackage{emnlp2017}
\usepackage{times}
\usepackage{latexsym}
\usepackage{booktabs}
\usepackage{todonotes}
\usepackage{blindtext}
\usepackage[font={small}]{caption}
\usepackage{basecommon}
\usepackage{multirow}
\PassOptionsToPackage{hyphens}{url}
\usepackage[hyphens]{url}
\usepackage[anythingbreaks]{breakurl}

\emnlpfinalcopy

\def\emnlppaperid{***}

\title{Challenges in Data-to-Document Generation}

 \author{Sam Wiseman \and Stuart M. Shieber \and Alexander M. Rush \\
         School of Engineering and Applied Sciences \\ Harvard University \\ Cambridge, MA, USA \\ {\tt \{swiseman,shieber,srush\}@seas.harvard.edu }}

\date{}

\begin{document}

\maketitle

\begin{abstract}

  Recent neural models have shown significant progress on the problem of
  generating short descriptive texts conditioned on a small number of database records. In this work, we 
  suggest a slightly more difficult data-to-text generation task, and investigate how effective current approaches are on this task. In particular, we introduce a
  new, large-scale corpus of data records paired with descriptive
  documents, propose a series of extractive evaluation methods for
  analyzing performance, and obtain baseline results using current neural generation methods. Experiments show that these models produce
  fluent text, but fail to convincingly approximate human-generated documents. Moreover, even templated baselines exceed the performance of these neural models on some metrics, though copy- and reconstruction-based extensions lead to noticeable
  improvements. 
\end{abstract}

\section{Introduction}
Over the past several years, neural text generation systems have shown impressive performance on tasks such as machine translation and summarization. As neural systems begin to move toward generating longer outputs in response to longer and more complicated inputs, however, the generated texts begin to display reference errors, inter-sentence incoherence, and a lack of fidelity to the source material. The goal of this paper is to suggest a particular, long-form generation task in which these challenges may be fruitfully explored, to provide a publically available dataset for this task, to suggest some automatic evaluation metrics, and finally to establish how current, neural text generation methods perform on this task.

A classic problem in natural-language generation (NLG)
\cite{kukich1983design,mckeown1992text,reiter1997building} involves taking
structured data, such as a table, as input, and producing text that adequately and fluently describes this data as output. Unlike machine translation, which aims for a
complete transduction of the sentence to be translated, this form of NLG is typically taken to require addressing (at least) two separate challenges: \textit{what to say}, the selection of an appropriate subset of the input data to discuss, and \textit{how to say it}, the surface realization of a generation~\citep{reiter1997building,jurafsky2014speech}. Traditionally, these two challenges have been 
modularized and handled separately by generation systems. However, neural generation systems, which are typically trained end-to-end as conditional language models~\cite{mikolov-2010,sutskever2011generating,sutskever2014sequence}, blur this distinction. 

In this context, we believe the problem of generating multi-sentence summaries of tables or database records to be a reasonable next-problem for neural techniques to tackle as they begin to consider more difficult NLG tasks. In particular, we would like this generation task to have the following two properties: (1) it is relatively easy to obtain fairly clean summaries and their corresponding databases for dataset construction, and (2) the summaries should be primarily focused on conveying the information in the database. This latter property ensures that the task is somewhat congenial to a standard encoder-decoder approach, and, more importantly, that it is reasonable to \textit{evaluate} generations in terms of their fidelity to the database.

One task that meets these criteria is that of generating summaries of sports games from associated box-score data, and there is indeed a long history of NLG work that generates sports game summaries~\citep{robin1994revision,tanaka1998reactive,
barzilay2005collective}. To this end, we make the following contributions:
\begin{itemize}
\item We introduce a new large-scale corpus
consisting of textual descriptions of basketball games paired with
extensive statistical tables. This dataset is sufficiently large that fully data-driven
approaches might be sufficient. 
\item We
introduce a series of extractive evaluation models to automatically
evaluate output generation performance, exploiting the fact that post-hoc information extraction is significantly easier than generation itself. 
\item  We apply a series of state-of-the-art neural methods, as well as a simple templated generation system, to our
data-to-document generation task in order to establish baselines and study
their generations.
\end{itemize}

Our experiments indicate that neural systems are
quite good at producing fluent outputs and generally score well on standard
word-match metrics, but perform quite poorly at content selection and at capturing long-term structure. While the use of copy-based models and additional reconstruction terms in the training loss can lead to improvements in
BLEU and in our proposed extractive evaluations, current models are still
quite far from producing human-level output, and are significantly
worse than templated systems in terms of content selection and realization. Overall, we believe this
problem of data-to-document generation highlights important remaining
challenges in neural generation systems, and the use of extractive
evaluation reveals significant issues hidden by standard automatic
metrics.

\begin{figure*}
\centering
\scalebox{0.95}{
\begin{tikzpicture}

\node[draw, inner sep=10pt, rounded corners, text width=25em, xshift=-8.5cm,]{\small
  \begin{center}

\begin{tabular}{lcccccc}
\toprule
{} & WIN & LOSS & PTS & FG\_PCT & RB & AS \ldots \\
TEAM &           &             &          &             &          &          \\
\midrule
Heat      &        11 &          12 &      103 &          49 &       47 &       27 \\
Hawks     &         7 &          15 &       95 &          43 &       33 &       20 \\
\bottomrule
\end{tabular}
\vspace{0.5cm}

\begin{tabular}{lccccccccc}
\toprule
{} &  AS &    RB &   PT &  FG &  FGA & CITY  $\ldots$ \\
PLAYER      &      &      &      &       &      &      &           \\
\midrule
Tyler Johnson    &    5 &    2 &  27 &    8 &   16 &     Miami \\
Dwight Howard    &    4 &    17 &  23 &    9 &   11 &   Atlanta \\
Paul Millsap     &    2 &    9 &  21 &    8 &   12 &   Atlanta \\
Goran Dragic     &    4 &    2 &  21 &    8 &   17 &     Miami \\
Wayne Ellington  &    2 &    3 &  19 &    7 &   15 &     Miami \\
Dennis Schroder  &    7 &    4 &  17 &    8 &   15 &   Atlanta \\
Rodney McGruder  &    5 &    5 &  11 &    3 &    8 &     Miami \\
Thabo Sefolosha  &    5 &    5 &  10 &    5 &   11 &   Atlanta \\
Kyle Korver      &    5 &    3 &  9 &    3 &    9 &   Atlanta \\
\ldots \\
\bottomrule
\end{tabular}
  \end{center}

};

\node [rectangle, draw,thick,fill=blue!0,text width=16em,  rounded corners, inner sep =10pt, minimum height=1em]{\baselineskip=100pt \small  The Atlanta Hawks defeated the Miami Heat , 103 - 95 , at Philips Arena on Wednesday . Atlanta was in desperate need of a win and they were able to take care of a shorthanded Miami team here . Defense was key for the Hawks , as they held the Heat to 42 percent shooting and forced them to commit 16 turnovers . Atlanta also dominated in the paint , winning the rebounding battle , 47 - 34 , and outscoring them in the paint 58 - 26.The Hawks shot 49 percent from the field and assisted on 27 of their 43 made baskets . This was a near wire - to - wire win for the Hawks , as Miami held just one lead in the first five minutes . Miami ( 7 - 15 ) are as beat - up as anyone right now and it 's taking a toll on the heavily used starters . Hassan Whiteside really struggled in this game , as he amassed eight points , 12 rebounds and one blocks on 4 - of - 12 shooting ... \par};
\end{tikzpicture}
}
\caption{An example data-record and document pair from the \textsc{RotoWire}
  dataset. We show a subset of the game's records (there
  are 628 in total), and a selection from the gold document. The document mentions only a select subset of the records, but may
  express them in a complicated manner. In
  addition to capturing the writing style, a generation system should select similar record content, express it clearly, and order it
  appropriately.
}
\label{fig:samplesummary}
\end{figure*}

\section{Data-to-Text Datasets}

We consider the problem of generating descriptive text from database
records. Following the
notation in \citet{liang2009learning}, let $\bolds = \{r_j\}_{j=1}^J$
be a set of records, where for each $r \, {\in} \, \bolds$ we define
$r.t \, {\in} \, \mcT$ to be the \textit{type} of $r$, and we assume each $r$ to be a binarized
relation, where $r.e$ and $r.m$ are a record's entity
and value, respectively. For example, a database recording statistics for a basketball game might have a record $r$ such that $r.t$ =
\textsc{points}, $r.e$ = \textsc{Russell Westbrook}, and $r.m=50$. In this case, $r.e$ gives the
player in question, and $r.m$ gives the
number of points the player scored. From these records, we are interested in generating descriptive text, $\hat{y}_{1:T} = \hat{y}_1, \ldots, \hat{y}_T$ of $T$ words such that $\hat{y}_{1:T}$ is an adequate and fluent summary of $\bolds$. A dataset for training data-to-document systems typically consists of $(\bolds, y_{1:T})$ pairs, where $y_{1:T}$ is a document consisting of a gold (i.e., human generated) summary for database $\bolds$.

Several benchmark datasets have been used in recent years for the text
generation task, the most popular of these being
\textsc{WeatherGov}~\cite{liang2009learning} and
\textsc{Robocup}~\cite{chen2008learning}. Recently, neural generation
systems have show strong results on these datasets, with the system of
\newcite{mei2016what} achieving BLEU scores in the 60s and 70s on
\textsc{WeatherGov}, and BLEU scores of almost 30 even on the smaller
\textsc{Robocup} dataset. These results are quite promising, and suggest that neural models are
a good fit for text generation. However, the statistics of these
datasets, shown in Table~\ref{tab:datastats}, indicate that these
datasets use relatively simple language and record structure. Furthermore, there is reason to believe that \textsc{WeatherGov} is at least partially machine-generated~\citep{reiterblog}.
More recently, \citet{lebret2016neural} introduced the \textsc{WikiBio} dataset, which is at least an order of magnitude larger in terms of number of tokens and record types. However, as shown in Table~\ref{tab:datastats}, this dataset too only contains short (single-sentence) generations, and relatively few records per generation. As such, we
believe that early success on these datasets is not yet sufficient for
testing the desired linguistic capabilities of text generation at a
document-scale.

With this challenge in mind, we introduce a new dataset for data-to-document text generation, available at \url{https://github.com/harvardnlp/boxscore-data}. The dataset is intended
to be comparable to \textsc{WeatherGov} in terms of token count, but to have significantly longer target texts, a larger vocabulary
space, and to require more difficult content selection.

The dataset consists of two sources of articles summarizing NBA basketball games, paired with their corresponding box- and line-score tables. The data statistics of these two sources, \textsc{RotoWire} and \textsc{SBNation}, are also shown in Table~\ref{tab:datastats}. The first dataset, \textsc{RotoWire}, uses professionally written, medium length game summaries targeted at fantasy basketball fans. The writing is colloquial, but relatively well structured, and targets an audience primarily interested in game statistics.
The second dataset, \textsc{SBNation}, uses fan-written summaries targeted at other fans. This dataset is significantly larger, but also much more challenging, as the language is very informal, and often tangential to the statistics themselves.
We show some sample text from \textsc{RotoWire} in Figure~\ref{fig:samplesummary}. 
Our primary focus will be on the \textsc{RotoWire} data.

\begin{table}[t]
\centering
\small
\begin{tabular}{llllll}
\toprule
& \textsc{RC} &\textsc{WG} & \textsc{WB} & \textsc{RW}   & \textsc{SBN}\\
\midrule
Vocab   & 409   & 394 & 400K & 11.3K  & 68.6K \\
Tokens &  11K  & 0.9M  & 19M & 1.6M & 8.8M \\
Examples & 1.9K & 22.1K & 728K & 4.9K  & 10.9K \\
Avg Len & 5.7  &  28.7 & 26.1 & 337.1 & 805.4 \\   
Rec. Types & 4 & 10 & 1.7K & 39 & 39\\
Avg Records & 2.2 & 191 & 19.7 & 628 & 628\\
\bottomrule
\end{tabular}
\caption{Vocabulary size, number of total tokens, number of distinct examples, average generation length, total number of record types, and average number of records per example for the \textsc{Robocup} (RC), \textsc{WeatherGov} (WG), \textsc{WikiBio} (WB), \textsc{RotoWire} (RW), and \textsc{SBNation} (SBN) datasets.}
\label{tab:datastats}
\end{table}

\section{Evaluating Document Generation}
We begin by discussing the evaluation of generated documents, since both the task we introduce and the evaluation methods we propose are motivated by some of the shortcomings of current approaches to evaluation. Text generation systems are typically evaluated using a combination
of automatic measures, such as BLEU~\cite{papineni02bleu}, and human evaluation. While BLEU is perhaps a reasonably effective way of evaluating short-form text generation, we found it to be unsatisfactory
for document generation. In particular, we note that it primarily rewards fluent
text generation, rather than generations that capture the most important information in the database, or that report the information in a particularly coherent way. While human evaluation, on the other hand, is likely ultimately necessary for evaluating generations~\cite{liu2016how,wu2016google}, it is much less convenient than using automatic metrics. Furthermore, we believe that current text generations are sufficiently bad in sufficiently obvious ways that automatic metrics can still be of use in evaluation, and we are not yet at the point of needing to rely solely on human evaluators.

\subsection{Extractive Evaluation}
\label{sec:extractive}
To address this evaluation challenge, we begin with the intuition that assessing document quality is easier than document generation. In particular, it is much easier to automatically extract information from documents than to generate documents that accurately convey desired information. As such, simple, high-precision information extraction models can serve
as the basis for assessing and better understanding the quality of automatic generations. 
We emphasize that such an evaluation scheme is most appropriate when evaluating generations (such as basketball game summaries) that are primarily intended to summarize information. While many generation problems do not fall into this category, we believe this to be an interesting category, and one worth focusing on \textit{because} it is amenable to this sort of evaluation.

To see how a simple information extraction system might work, consider the document in Figure~\ref{fig:samplesummary}. We may first extract candidate entity (player, team, and city) and value (number and
certain string) pairs $r.e, r.m$ that appear in the text, and then predict the type $r.t$ (or none) of each candidate pair. For example, we might extract the entity-value pair (``Miami Heat'', ``95'') from the first sentence in Figure~\ref{fig:samplesummary}, and then predict that the \textit{type} of this pair is \textsc{points}, giving us an extracted record $r$ such that $(r.e, r.m, r.t) = $ (\textsc{Miami Heat}, 95, \textsc{points}). Indeed, many relation extraction systems
reduce relation extraction to multi-class classification precisely in this
way~\citep{zhang2004weakly,zhou2008semi,zeng2014relation,dosSantos2015classifying}. 

More concretely, given a document $\hat{y}_{1:T}$, we consider all pairs of word-spans in each sentence
that represent possible entities $e$ and values $m$. We then model
$p(r.t \, | \, e, m;\btheta)$ for each pair, using $r.t =\epsilon$ to
indicate unrelated pairs. We use architectures similar to those discussed in \citet{collobert2011natural} and
\citet{dosSantos2015classifying} to parameterize this probability; full details are given in the Appendix. 

Importantly, we note that the $(\bolds, y_{1:T})$ pairs typically used for training data-to-document systems are also sufficient for training the information
extraction model presented above, since we can obtain (partial) supervision by simply checking whether
a candidate record lexically matches a record in $\bolds$.\footnote{Alternative approaches
  explicitly align the document with the table for this task \cite{liang2009learning}.} However, since there may be multiple records $r \, {\in} \, \bolds$ with the same $e$ and $m$ but with different types $r.t$, we will not always be able to determine the type of a given entity-value pair found in the text. We therefore train our classifier to
minimize a latent-variable loss: for all document spans $e$ and $m$, with observed types $t(e,m)= \{r.t : r \, {\in} \, \bolds, r.e \, {=} \, e, r.m \, {=} \, m\}$ (possibly $\{\epsilon\}$), we minimize
\begin{align*}
\mcL(\btheta) = - \sum_{e,m} \log \sum_{t'\in t(e,m)} p(r.t=t' \given e,m ;\btheta).
\end{align*} 

\noindent We find that this simple system trained in this way is
quite accurate at predicting relations. On the \textsc{Rotowire}
data it achieves over 90\% accuracy on held-out data, and recalls approximately 60\% of the relations licensed by the records.

\subsection{Comparing Generations} 
\label{sec:comparing}
With a sufficiently precise relation extraction system, we can begin to evaluate how well an automatic generation $\hat{y}_{1:T}$ has captured the information in a set of records $\bolds$. In particular, since the predictions of a precise information extraction system serve to align entity-mention pairs in the text with database records, this alignment can be used both to evaluate a generation's content selection (``what the generation says''), as well as content placement (``how the generation says it'').

We consider in particular three induced metrics:
\begin{itemize}
\item Content Selection (CS): precision and recall of unique relations $r$ extracted from $\hat{y}_{1:T}$ that are also extracted from $y_{1:T}$. This measures how well the generated document matches the gold document in terms of selecting which records to generate. 
\item Relation Generation (RG): precision and number of unique relations $r$ extracted from $\hat{y}_{1:T}$ that also appear in $\bolds$. This measures how well the system is able to generate text containing factual (i.e., correct) records. 
\item Content Ordering (CO): normalized Damerau-Levenshtein
  Distance~\cite{brill2000improved}\footnote{DLD is a variant of
    Levenshtein distance that allows transpositions of elements; it is useful in comparing the
    ordering of sequences that may not be permutations of the same set (which is a requirement for measures like Kendall's Tau). }
  between the sequences of records extracted from $y_{1:T}$ and that extracted
  from $\hat{y}_{1:T}$. This measures how well the system orders the records it chooses to discuss.
\end{itemize} 

\noindent We note that CS primarily targets the ``what to say'' aspect of evaluation, CO targets the ``how to say it'' aspect, and RG targets both. 

We conclude this section by contrasting the automatic evaluation we have proposed with recently proposed \textit{adversarial evaluation} approaches, which also advocate automatic metrics backed by classification~\cite{bowman2016generating,kannan2016adversarial, li2017adversarial}. Unlike adversarial evaluation, which uses a black-box classifier to determine the quality of a generation, our metrics are defined with respect to the predictions of an information extraction system. Accordingly, our metrics are quite interpretable, since by construction it is always possible to determine which fact (i.e., entity-value pair) in the generation is determined by the extractor to not match the database or the gold generation.

\section{Neural Data-to-Document Models}
\label{sec:models}
In this section we briefly describe the neural generation methods we apply to the proposed task. As a base model we utilize the now standard attention-based encoder-decoder model~\citep{sutskever2014sequence,cho2014on,bahdanau2015neural}. We also experiment with several recent extensions to this model, including copy-based generation, and training with a source reconstruction term in the loss (in addition to the standard per-target-word loss). 

\paragraph{Base Model}

For our base model, we map each record $r \, {\in} \, \bolds$ into a vector $\tilde{\boldr}$  by first embedding $r.t$ (e.g., \textsc{points}), $r.e$ (e.g., \textsc{Russell Westbrook}), and $r.m$ (e.g., 50), and then applying a 1-layer MLP (similar to~\citet{yang2016reference}).\footnote{We also include an additional feature for whether the player is on the home- or away-team.} Our source data-records are then represented as $\tilde{\bolds} = \{\tilde{\boldr}_j\}_{j=1}^J$. Given $\tilde{\bolds}$, we use an LSTM decoder with attention and input-feeding, in the style of \citet{luong2015effective}, to compute the probability of each target word, conditioned on the previous words and on $\bolds$. The model is trained end-to-end to minimize the negative log-likelihood of the words in the gold text $y_{1:T}$ given corresponding source material $\bolds$.

\paragraph{Copying}

There has been a surge of recent work involving augmenting encoder-decoder models to copy words directly from the source material on which they condition~\cite{gu2016incorporating,gulcehre2016pointing,
merity2016pointer,jia2016data,yang2016reference}.
These  models typically introduce an additional binary variable $z_t$ into the per-timestep target word distribution, which indicates whether the target word $\hat{y}_t$ is copied from the source or generated: 
\begin{align*} 
p(\hat{y}_t \given \hat{y}_{1:t-1}, \bolds) = \sum_{z \in \{0,1\}} p(\hat{y}_t, z_t = z \given \hat{y}_{1:t-1}, \bolds).
\end{align*}
In our case, we assume that target words are copied from the \textit{value} portion of a record $r$; that is, a copy implies $\hat{y}_t \, {=} \, r.m$ for some $r$ and $t$.

\paragraph{Joint Copy Model} The models of \citet{gu2016incorporating} and \citet{yang2016reference} parameterize the \textit{joint} distribution table over $\hat{y}_t$ and $z_t$ directly:
\begin{align*}
p(&\hat{y}_t, z_t \given \hat{y}_{1:t-1}, \bolds) \propto \\ 
&\begin{cases} \mathrm{copy}( \hat{y}_{t}, \hat{y}_{1:t-1}, \bolds) &\mbox{$z_t = 1$, $\hat{y}_t \in \bolds$} \\
0 &\mbox{$z_t = 1$, $\hat{y}_t \not \in \bolds$} \\
\mathrm{gen}( \hat{y}_{t}, \hat{y}_{1:t-1}, \bolds) &\mbox{$z_t=0$}, \end{cases}
\end{align*}
where $\mathrm{copy}$ and $\mathrm{gen}$ are functions parameterized in terms of the decoder RNN's hidden state that assign scores to words, and where the notation $\hat{y}_t \, {\in} \, \bolds$ indicates that $\hat{y}_t$ is equal to $r.m$ for some $r \, {\in} \, \bolds$.

\paragraph{Conditional Copy Model} \citet{gulcehre2016pointing}, on the other hand, decompose the joint probability as:
\begin{align*}
&p(\hat{y}_t, z_t \given \hat{y}_{1:t-1}, \bolds) = \\ &\begin{cases} p_{\mathrm{copy}}(\hat{y}_t \given z_t, \hat{y}_{1:t-1}, \bolds) \, p(z_t \given \hat{y}_{1:t-1}, \bolds) &\mbox{$z_t{=}1$} \\
p_{\mathrm{gen}}(\hat{y}_t \given z_t, \hat{y}_{1:t-1}, \bolds) \, p(z_t \given \hat{y}_{1:t-1}, \bolds) &\mbox{$z_t{=}0$}, \end{cases}
\end{align*}

\noindent where an MLP is used to model $p(z_t \given \hat{y}_{1:t-1}, \bolds)$. 

Models with copy-decoders may be trained to minimize the negative log marginal probability, marginalizing out the latent-variable $z_t$~\cite{gu2016incorporating,
yang2016reference,merity2016pointer}. However, if it is known which target words $y_t$ are copied, it is possible to train with a loss that does not marginalize out the latent $z_t$. \citet{gulcehre2016pointing}, for instance, assume that any target word $y_t$ that also appears in the source is copied, and train to minimize the negative joint log-likelihood of the $y_t$ and $z_t$. 

In applying such a loss in our case, we again note that there may be multiple records $r$ such that $r.m$ appears in $\hat{y}_{1:T}$. Accordingly, we slightly modify the $p_{\mathrm{copy}}$ portion of the loss of \citet{gulcehre2016pointing} to sum over all matched records. In particular, we model the probability of relations $r \in \bolds$ such
that $r.m = y_t$ and $r.e$ is in the same sentence as $r.m$. Letting $r(y_t) = \{ {r\in \bolds: r.m=y_t, \mathrm{same{-}sentence}(r.e, r.m)}  \}$, we have:
\begin{align*}
p_{\text{copy}}(y_t \given z_t,  y_{1:t-1}, \bolds) &= \sum_{r \in r(y_t)} p(r \given z_t,  y_{1:t-1}, \bolds).
\end{align*} 

We note here that the key distinction for our purposes between the Joint Copy model and the Conditional Copy model is that the latter \textit{conditions} on whether there is a copy or not, and so in $p_{\mathrm{copy}}$ the source records compete only with each other. In the Joint Copy model, however, the source records also compete with words that cannot be copied. As a result, training the Conditional Copy model with the supervised loss of \citet{gulcehre2016pointing} can be seen as training with a word-level reconstruction loss, where the decoder is trained to choose the record in $\bolds$ that gives rise to $y_t$.

\paragraph{Reconstruction Losses}

Reconstruction-based techniques can also be applied at the
document- or sentence-level during training. One simple approach to this problem is to
utilize the hidden states of the decoder to try to reconstruct the
database. A fully differentiable approach using the decoder hidden states has recently been
successfully applied to neural machine translation
by~\citet{tu2017neural}. Unlike copying, this method is applied only
at training, and attempts to learn decoder hidden states with broader 
coverage of the input data.

In adopting this reconstruction approach we 
segment the decoder hidden states $\boldh_t$ into
$\lceil \frac{T}{B} \rceil$ contiguous blocks of size at most
$B$. Denoting a single one of these hidden state blocks as $\boldb_i$,
we attempt to predict each field value in some record
$r \in \bolds$ from $\boldb_i$. We define
$p(r.e, r.m \given \boldb_i)$, the probability of the entity and value in record $r$ given $\boldb_i$, to be
$\mathrm{softmax}(f(\boldb_i))$, where $f$ is a parameterized function of
$\boldb_i$, which in our experiments utilize a convolutional layer
followed by an MLP; full details are given in the Appendix. We further extend this
idea and predict $K$ records in $\bolds$ from $\boldb_i$,
rather than one. We can train with the following reconstruction loss for a particular $\boldb_i$:
\begin{align*}
\mcL(\btheta) &= -\sum_{k=1}^K \min_{r \in \bolds}  \log p_k(r \given \boldb_i; \btheta) \\
&= -\sum_{k=1}^K \min_{r \in \bolds}  \sum_{x \in \{e, m, t\}} \log p_k(r.x \given \boldb_i; \btheta) ,
\end{align*}
where $p_k$ is the $k$'th predicted distribution over records, and where we have modeled each component of $r$ independently. This loss attempts to make the \textit{most} probable record in $\bolds$ given $\boldb_i$ more probable. We found that augmenting the above loss with a term that penalizes the total variation distance (TVD) between the $p_k$ to be helpful.\footnote{Penalizing the TVD between the $p_k$ might be useful if, for instance, $K$ is too large, and only a smaller number of records can be predicted from $\boldb_i$. We also experimented with \textit{encouraging}, rather than penalizing the TVD between the $p_k$, which might make sense if we were worried about ensuring the $p_k$ captured different records. 
} Both $\mcL(\btheta)$ and the TVD term are simply added to the standard negative log-likelihood objective at training time.

\section{Experimental Methods}

In this section we highlight a few important details of our models and methods; full details are in the Appendix. For our \textsc{RotoWire} models, the record encoder produces $\tilde{\boldr}_j$ in $\reals^{600}$, and we use a 2-layer LSTM
decoder with hidden states of the same size as the $\tilde{\boldr}_j$,
and dot-product attention and input-feeding in the style of
\citet{luong2015effective}. Unlike past work, we use two identically structured
attention layers, one to compute the standard generation probabilities ($\mathrm{gen}$ or $p_{\mathrm{gen}}$), and one to produce the scores used in 
$\mathrm{copy}$ or $p_{\mathrm{copy}}$. 

We train the generation models using SGD and
truncated BPTT~\cite{elman1990,mikolov-2010}, as in language
modeling. That is, we split each $y_{1:T}$ into contiguous blocks of length 100, and backprop both the gradients with respect to the current block as well as with respect to the encoder parameters for each block.

Our extractive evaluator consists of an ensemble of 3 single-layer
convolutional and 3 single-layer bidirectional LSTM models. The convolutional models
concatenate convolutions with kernel widths 2, 3, and 5, and 200
feature maps in the style of \cite{kim2014convolutional}.
Both models are trained with SGD.

\paragraph{Templatized Generator} In addition to neural baselines, we also use a problem-specific,
template-based generator. The template-based generator first emits a sentence about the teams playing in the game, using a templatized
sentence taken from the training set:

\begin{quote}
{\small The \texttt{<team1>} (\texttt{<wins1>}-\texttt{<losses1>}) defeated the \texttt{<team2>} (\texttt{<wins2>}-\texttt{<losses2>}) \texttt{<pts1>}-\texttt{<pts2>}. }
\end{quote}

\noindent Then, 6 player-specific sentences of the following form are emitted (again adapting a simple sentence from the training set):
\begin{quote}
{\small \texttt{<player>} scored \texttt{<pts>} points (\texttt{<fgm>}-\texttt{<fga>} FG, \texttt{<tpm>}-\texttt{<tpa>} 3PT, \texttt{<ftm>}-\texttt{<fta>} FT) to go with \texttt{<reb>} rebounds. }
\end{quote}

\noindent The 6 highest-scoring players in the game are used to fill in the above template. Finally, a typical end sentence is emitted:

\begin{quote}
{\small The \texttt{<team1>}' next game will be at home against the Dallas Mavericks, while the \texttt{<team2>} will travel to play the Bulls.}
\end{quote}

Code implementing all models can be found at \url{https://github.com/harvardnlp/data2text}. Our encoder-decoder models are based on OpenNMT~\cite{opennmt}.

\section{Results}
We found that all models performed quite poorly on the \textsc{SBNation} data, with the best model achieving a validation perplexity of 33.34 and a BLEU score of 1.78. This poor performance is presumably attributable to the noisy quality of the \textsc{SBNation} data, and the fact that many documents in the dataset focus on information not in the box- and line-scores. Accordingly, we focus on \textsc{RotoWire} in what follows. 

The main results for the \textsc{RotoWire} dataset are shown in Table~\ref{tab:maintestresults}, which shows the performance of the models in Section~\ref{sec:models} in terms of the metrics defined in Section~\ref{sec:comparing}, as well as in terms of perplexity and BLEU. 

\subsection{Discussion}
\begin{table*}
\small
\centering
\begin{tabular}{llccccccc}
\toprule
  & & \multicolumn{7}{c}{Development} \\
\midrule
& & \multicolumn{2}{c}{RG}  & \multicolumn{2}{c}{CS} & CO & PPL & BLEU\\
Beam & Model & P\% & \# & P\% & R\% & DLD\% &  & \\
\midrule
&Gold                 & 91.77 & 12.84 & 100   & 100   & 100  & 1.00 & 100 \\
&Template             & 99.35 & 49.7  & 18.28 & 65.52 & 12.2 & N/A  & 6.87   \\
\midrule
\multirow{ 4}{*}{B=1} & Joint Copy             & 47.55 & 7.53  & 20.53 & 22.49 & 8.28 & 7.46 & 10.41 \\

&Joint Copy + Rec  & 57.81 & 8.31  & 23.65 & 23.30 & 9.02 & 7.25 & 10.00 \\

&Joint Copy  + Rec + TVD  & 60.69 & 8.95 & 23.63 & 24.10 & 8.84 & 7.22 & 12.78 \\

&Conditional Copy       & 68.94 & 9.09  & 25.15 & 22.94 & 9.00 & 7.44 & 13.31 \\
\midrule
\multirow{ 4}{*}{B=5}  &Joint Copy             & 47.00 & 10.67 & 16.52 & 26.08 & 7.28 & 7.46 & 10.23 \\
&Joint Copy  + Rec  & 62.11 & 10.90 & 21.36 & 26.26 & 9.07 & 7.25 & 10.85 \\
&Joint Copy  + Rec + TVD  & 57.51 & 11.41 & 18.28 & 25.27 & 8.05 & 7.22 & 12.04 \\
&Conditional Copy        & 71.07 & 12.61 & 21.90 & 27.27 & 8.70 & 7.44 & 14.46 \\
\midrule

  & & \multicolumn{7}{c}{Test} \\
\midrule
&Template                 & 99.30 & 49.61 & 18.50 & 64.70 & 8.04 & N/A  & 6.78   \\
&Joint Copy + Rec (B=5)      & 61.23 & 11.02 & 21.56 & 26.45 & 9.06 & 7.47 & 10.88 \\
&Joint Copy + Rec + TVD (B=1) & 60.27 & 9.18  & 23.11 & 23.69 & 8.48 & 7.42 & 12.96 \\
&Conditional Copy (B=5)            & 71.82 & 12.82 & 22.17 & 27.16 & 8.68 & 7.67 & 14.49 \\
\bottomrule
\end{tabular}
\caption{Performance of induced metrics on gold and system outputs of RotoWire development and test data. 
  Columns indicate Record Generation (RG) precision and count, Content Selection (CS) precision and recall, 
  Count Ordering (CO) in normalized Damerau-Levenshtein distance, perplexity, and BLEU. These first three metrics are described in Section~\ref{sec:comparing}. Models compare Joint and Conditional 
  Copy also with addition Reconstruction loss and Total Variation Distance extensions (described in Section~\ref{sec:models}).
}
\label{tab:maintestresults}
\end{table*} 
 
There are several interesting relationships in the development portion of Table~\ref{tab:maintestresults}.
First we note that the Template model scores very poorly on BLEU, but
does quite well on the extractive metrics, providing an upper-bound
for how domain knowledge could help content selection and
generation. All the neural models make significant improvements in terms
of BLEU score, with the conditional copying with beam search
performing the best, even though all the neural models achieve roughly
the same perplexity.

The extractive metrics provide further insight into the behavior of
the models. We first note that on the gold documents $y_{1:T}$, the extractive model reaches
$92\%$ precision. Using the Joint Copy model, generation only
has a record generation (RG) precision of $47\%$ indicating that
relationships are often generated incorrectly. The best Conditional
Copy system improves this value to $71\%$, a significant improvement
and potentially the cause of the improved BLEU score, but still far below
gold. 

Notably, content selection (CS) and content ordering (CO) seem to have
no correlation at all with BLEU. There is some improvement with CS for
the conditional model or reconstruction loss, but not much change as
we move to beam search. CO actually gets worse as beam search is
utilized, possibly a side effect of generating more records (RG\#). 
The fact that these scores are much worse than the simple templated model
indicates that further research is needed into better copying alone for 
content selection and better long term content ordering models. 

Test results are consistent with development results, indicating that the Conditional Copy model is most 
effective at BLEU, RG, and CS, and that reconstruction is quite helpful for 
improving the joint model.

\begin{figure}[t]
  \small
  \scalebox{0.9}{
  \begin{tikzpicture}
    
    \node[text width=8cm, draw, rounded corners, inner sep=10pt]{\small
  The \textcolor{blue}{Utah Jazz} ( \textcolor{blue}{38} - \textcolor{red}{26} ) \textcolor{red}{defeated} the \textcolor{blue}{Houston Rockets} ( \textcolor{red}{38} - \textcolor{blue}{26} ) \textcolor{blue}{117 - 91} on Wednesday at \textcolor{blue}{Energy Solutions Arena in Salt Lake City} . The \textcolor{blue}{Jazz} got out to a quick start in this one , out - scoring the \textcolor{blue}{Rockets} \textcolor{blue}{31} - \textcolor{red}{15} in the first quarter alone . Along with the quick start , the \textcolor{blue}{Rockets} were the superior shooters in this game , going \textcolor{blue}{54 percent from the field} and \textcolor{blue}{43 percent from the three - point line} , while the \textcolor{blue}{Jazz} went \textcolor{blue}{38} percent from the floor and a meager \textcolor{blue}{19} percent from deep . The \textcolor{blue}{Rockets} were able to out - rebound the \textcolor{red}{Rockets} \textcolor{blue}{49} - \textcolor{red}{49} , giving them just enough of an advantage to secure the victory in front of their \textcolor{red}{home crowd} . The \textcolor{blue}{Jazz} were led by the duo of \textcolor{blue}{Derrick Favors} and \textcolor{red}{James Harden} . Favors went \textcolor{blue}{2 - for - 6} from the field and \textcolor{blue}{0 - for - 1} from the three - point line to score a \textcolor{red}{game - high} of \textcolor{red}{15 points} , while also adding \textcolor{blue}{four rebounds} and \textcolor{red}{four assists} ....};
  \end{tikzpicture}
  }
\caption{Example document generated by the Conditional Copy system with a beam of size 5. Text that accurately reflects a record in the associated box- or line-score is highlighted in \textcolor{blue}{blue}, and erroneous text is highlighted in \textcolor{red}{red}.}
\label{fig:qualitative}
\end{figure} 
 
\subsection{Human Evaluation}
We also undertook two human evaluation studies, using Amazon Mechanical Turk. The first study attempted to determine whether generations considered to be more precise by our metrics were also considered more precise by human raters. To accomplish this, raters were presented with a particular NBA game's box score and line score, as well as with (randomly selected) sentences from summaries generated by our different models for those games. Raters were then asked to count how many facts in each sentence were supported by records in the box or line scores, and how many were contradicted. We randomly selected 20 distinct games to present to raters, and a total of 20 generated sentences per game were evaluated by raters. The left two columns of Table~\ref{tab:humantable} contain the average numbers of supporting and contradicting facts per sentence as determined by the raters, for each model. We see that these results are generally in line with the RG and CS metrics, with the Conditional Copy model having the highest number of supporting facts, and the reconstruction terms significantly improving the Joint Copy models.

Using a Tukey HSD post-hoc analysis of an ANOVA with the number of contradicting facts as the dependent variable and the generating model and rater id as independent variables, we found significant ($p < 0.01$) pairwise differences in contradictory facts between the gold generations and all models except ``Copy+Rec+TVD,'' as well as a significant difference between ``Copy+Rec+TVD'' and ``Copy''. We similarly found a significant pairwise difference between ``Copy+Rec+TVD'' and ``Copy'' for number of supporting facts.

Our second study attempted to determine whether generated summaries differed in terms of how natural their \textit{ordering} of records (as captured, for instance, by the DLD metric) is. To test this, we presented raters with random summaries generated by our models and asked them to rate the naturalness of the ordering of facts in the summaries on a 1-7 Likert scale. 30 random summaries were used in this experiment, each rated 3 times by distinct raters. The average Likert ratings are shown in the rightmost column of Table~\ref{tab:humantable}. While it is encouraging that the gold summaries received a higher average score than the generated summaries (and that the reconstruction term again improved the Joint Copy model), a Tukey HSD analysis similar to the one presented above revealed no significant pairwise differences.

\begin{table}
\centering
\small
\begin{tabular}{l@{\hskip 1\tabcolsep}cc@{\hskip 1\tabcolsep}c}
\toprule
 &  \# Supp. & \# Cont. & Order Rat.  \\
\midrule
Gold             &  2.04  & 0.70 & 5.19 \\
Joint Copy             &  1.65  & 2.31 & 3.90 \\
Joint Copy + Rec       &  2.33  & 1.83 & 4.43 \\
Joint Copy + Rec +TVD  &  2.43  & 1.16 & 4.18 \\
Conditional Copy &  3.05  & 1.48 & 4.03 \\
\bottomrule
\end{tabular}
\caption{Average rater judgment of number of box score fields supporting (left column) or contradicting (middle column) a generated sentence, and average rater Likert rating for the naturalness of a summary's ordering (right column). All generations use B=1.}
\label{tab:humantable}
\end{table} 
 
\subsection{Qualitative Example}

Figure~\ref{fig:qualitative} shows a document generated by the Conditional Copy model, using a beam of size 5.  This particular generation evidently
has several nice properties: it nicely learns the colloquial
style of the text, correctly using idioms such as ``19 percent from
deep.'' It is also partially accurate in its use of the records; we highlight in blue when it generates text that is licensed by a record in the associated box- and line-scores.

At the same time, the generation also contains major logical errors. First, there are basic copying mistakes, such as flipping the teams' win/loss records. The system also makes obvious semantic errors; for instance, it generates the phrase ``the
Rockets were able to out-rebound the Rockets.'' Finally, we see the model hallucinates factual statements, such as ``in front of their home crowd,'' which is presumably likely according to the language model, but ultimately incorrect (and not supported by anything in the box- or line- scores).
In practice, our proposed extractive evaluation will pick up on many
errors in this passage. For instance, ``four assists'' is an RG error,
repeating the Rockets' rebounds could manifest in a lower CO score, and incorrectly indicating the win/loss records is a CS error.

\section{Related Work}
In this section we note additional related work not noted throughout. Natural language generation has been studied for decades~\citep{kukich1983design,mckeown1992text,reiter1997building}, and generating summaries of sports games has been a topic of interest for almost as long~\citep{robin1994revision,tanaka1998reactive,barzilay2005collective}.

Historically, research has focused on both content selection (``what to say'')~\citep{kukich1983design,mckeown1992text,
reiter1997building,duboue2003statistical,barzilay2005collective}, and surface realization (``how to say it'')~\citep{goldberg1994using,reiter2005choosing} with earlier work using (hand-built) grammars, and later work using SMT-like approaches~\citep{wong2007generation} or generating from PCFGs~\citep{belz2008automatic} or other formalisms~\citep{soricut2006stochastic,white2007towards}. In the late 2000s and early 2010s, a number of systems were proposed that did both~\citep{liang2009learning,angeli2010simple,
kim2010generative,lu2011probabilistic,konstas2013global}.

Within the world of neural text generation, some recent work has focused on conditioning language models on tables~\citep{yang2016reference}, and generating short biographies from Wikipedia Tables~\citep{lebret2016neural, chisholm2017learning}. \citet{mei2016what} use a neural encoder-decoder approach on standard record-based generation datasets, obtaining impressive results, and motivating the need for more challenging NLG problems. 

\section{Conclusion and Future Work}
This work explores the challenges facing neural data-to-document
generation by introducing a new dataset, and proposing
various metrics for automatically evaluating content selection,
generation, and ordering. We see that recent ideas in copying and
reconstruction lead to improvements on this task, but that there is 
a significant gap even between these neural models and templated systems. We hope to 
motivate researchers to focus further on generation
problems that are relevant both to content selection and surface realization, but may not be reflected clearly in the model's perplexity. 

Future work on this task might include approaches that process or attend to the source records in a more sophisticated way, generation models that attempt to incorporate semantic or reference-related constraints, and approaches to conditioning on facts or records that are not as explicit in the box- and line-scores.

\section*{Acknowledgments}
We gratefully acknowledge the support of a Google Research Award.

\bibliography{gen}
\bibliographystyle{emnlp_natbib}

\newpage
\section*{Appendix}
\subsection*{A. Additional Dataset Details}
The \textsc{RotoWire} data covers NBA games played between 1/1/2014 and 3/29/2017; some games have multiple summaries. The summaries have been randomly split into training, validation, and test sets consisting of 3398, 727, and 728 summaries, respectively.

The \textsc{SBNation} data covers NBA games played between 11/3/2006 and 3/26/2017; some games have multiple summaries. The summaries have been randomly split into training, validation, and test sets consisting of 7633, 1635, and 1635 summaries, respectively.

All numbers in the box- and line-scores (but not the summaries) are converted to integers; fractional numbers corresponding to percents are multiplied by 100 to obtain integers in $[0, 100]$. We show the \textit{types} of records in the data in Table~\ref{tab:types}.

\begin{table*}
\centering
\small
\begin{tabular}{cccccccccc}
\toprule
\textbf{Player Types} & \textsc{posn} & \textsc{min} & \textsc{pts} & \textsc{fgm} & \textsc{fga} & \textsc{fg-pct} & \textsc{fg3m} & \textsc{fg3a} & \textsc{fg3-pct}  \\
& \textsc{ftm} & \textsc{fta} & \textsc{ft-pct} & \textsc{oreb} & \textsc{dreb} & \textsc{reb} & \textsc{ast} & \textsc{tov} & \textsc{stl}  \\
& \textsc{blk} & \textsc{pf} & \textsc{full-name} & \textsc{name1} & \textsc{name2} & \textsc{city} & & & \\[2mm] 
\textbf{Team Types} & \textsc{pts-qtr1} & \textsc{pts-qtr2} & \textsc{pts-qtr3} & \textsc{pts-qtr4} & \textsc{pts} & \textsc{fg-pct} & \textsc{fg3-pct} & \textsc{ft-pct} & \textsc{reb} \\
& \textsc{ast} & \textsc{tov} & \textsc{wins} & \textsc{losses} & \textsc{city} & \textsc{name} & & & \\
\bottomrule
\end{tabular}
\caption{Possible Record Types}
\label{tab:types}
\end{table*}

\subsection*{B. Generation Model Details}
\label{sec:suppgen}
\paragraph{Encoder} For the \textsc{RotoWire} data, a relation $r$ is encoded into $\tilde{\boldr}$ by embedding each of $r.e$, $r.t$, $r.m$ and a ``home-or-away'' indicator feature in $\reals^{600}$, and applying a 1-layer MLP (with ReLU nonlinearity) to map the concatenation of these vectors back into $\reals^{600}$. To initialize the decoder LSTMs, we first mean-pool over the $\tilde{\boldr}_j$ by entity (giving one vector per entity), and then linearly transform the concatenation of these pooled entity-representations so that they can initialize the cells and hidden states of a 2-layer LSTM with states also in $\reals^{600}$. The \textsc{SBNation} setup is identical, except all vectors are in $\reals^{700}$.

\paragraph{Decoder} As mentioned in the body of the paper, we compute two different attention distributions (i.e., using different parameters) at each decoding step. For the Joint Copy model, one attention distribution is not normalized, and is normalized along with all the output-word probabilities. 

Within the Conditional Copy model we compute $p(z_t | \hat{y}_{1:t-1}, \bolds)$ by mean-pooling the $\tilde{\boldr}_j$, concatenating them with the current (topmost) hidden state of the LSTM, and then feeding this concatenation via a 1-layer ReLU MLP with hidden dimension 600, and with a Sigmoid output layer. 

For the reconstruction-loss, we feed blocks (of size at most 100) of the decoder's LSTM hidden states through a \cite{kim2014convolutional}-style convolutional model. We use kernels of width 3 and 5, 200 filters, a ReLU nonlinearity, and max-over-time pooling. To create the $p_k$, these now 400-dimensional features are then mapped via an MLP with a ReLU nonlinearity into 3 separate 200 dimensional vectors corresponding to the predicted relation's entity, value, and type, respectively. These 200 dimensional vectors are then fed through (separate) linear decoders and softmax layers in order to obtain distributions over entities, values, and types. We use $K=3$ distinct $p_k$.

Models are trained with SGD, a learning rate of 1 (which is divided by 2 every time validation perplexity fails to decrease), and a batch size of 16. We use dropout (at a rate of 0.5) between LSTM layers and before the linear decoder.

\subsection*{C. Information Extraction Details}
\paragraph{Data} To form an information extraction dataset, we first sentence-tokenize the gold summary documents $y_{1:T}$ using NLTK~\cite{bird2006nltk}. We then determine which word-spans $y_{i:j}$ could represent entities (by matching against players, teams, or cities in the database), and which word-spans $y_{k:l}$ could represent numbers (using the open source \texttt{text2num} library\footnote{https://github.com/exogen/text2num} to convert (strings of) number-words into numbers).\footnote{We ignore certain particularly misleading number-words, such as "three-point," where we should not expect a corresponding value of 3 among the records.} We then consider each $y_{i:j}, y_{k:l}$ pair in the same sentence, and if there is a record $r$ in the database such that $r.e \, {=} \, y_{i:j}$ and $r.m \, {=} \, \mathrm{text2num}(y_{k:l})$ we annotate the $y_{i:j}, y_{k:l}$ pair with the label $r.t$; otherwise, we give it a label of $\epsilon$.

\paragraph{Model}
We predict relations by ensembling 3 convolutional models and 3 bidirectional LSTM~\cite{hochreiter1997lstm,graves2005framewise} models. 
Each model consumes the words in the sentence, which are embedded in $\reals^{200}$, as well as the distances of each word in the sentence from both the entity-word-span and the number-word-spans (as described above), which are each embedded in $\reals^{100}$. These vectors are concatenated (into a vector in $\reals^{500}$) and fed into either a convolutional model or a bidirectional LSTM model. 

The convolutional model uses 600 total filters, with 200 filters for kernels of width 2, 3, and 5, respectively, a ReLU nonlinearity, and max-pooling. These features are then mapped via a 1-layer (ReLU) MLP into $\reals^{500}$, which predicts one of the 39 relation types (or $\epsilon$) with a linear decoder layer and softmax. 

The bidirectional LSTM model uses a single layer with 500 units in each direction, which are concatenated. The hidden states are max-pooled, and then mapped via a 1-layer (ReLU) MLP into $\reals^{700}$, which predicts one of the 39 relation types (or $\epsilon$) with a linear decoder layer and softmax.

\end{document}